# Human Attention during Goal-directed Reading Comprehension Relies on Task Optimization

Jiajie Zou[1], Yuran Zhang[1], Jialu Li[3], Xing Tian[3], Nai Ding[1,2*],

[1] Key Laboratory for Biomedical Engineering of Ministry of Education,

College of Biomedical Engineering and Instrument Sciences,

Zhejiang University, Hangzhou 310027, China

[2] Research Center for Applied Mathematics and Machine Intelligence,

Research Institute of Basic Theories, Zhejiang lab, Hangzhou 311121, China

[3] Division of Arts and Sciences, New York University Shanghai

**Short title:** Optimal Distribution of Reading Attention

**\*Corresponding author:**

Nai Ding,

Email: ding_nai@zju.edu.cn

College of Biomedical Engineering and Instrument Sciences,

Zhejiang University, Hangzhou 310027, China

**Abstract**

The computational principles underlying attention allocation in complex goal-directed tasks remain elusive. Goal-directed reading, i.e., reading a passage to answer a question in mind, is a common real-world task that strongly engages attention. Here, we investigate what computational models can explain attention distribution in this complex task. We show that the reading time on each word is predicted by the attention weights in transformer-based deep neural networks (DNNs) optimized to perform the same reading task. Eye-tracking further reveals that readers separately attend to basic text features and question-relevant information during first-pass reading and rereading, respectively. Similarly, text features and question relevance separately modulate attention weights in shallow and deep DNN layers. Furthermore, when readers scan a passage without a question in mind, their reading time is predicted by DNNs optimized for a word prediction task. Therefore, attention during real-world reading can be interpreted as the consequence of task optimization.

**Introduction**

Attention profoundly influences information processing in the brain [1, 2], and a large number of studies have been devoted to studying the neural mechanisms of attention. From the perspective of David Marr, the attention mechanism can be studied from three levels, i.e., the computational, algorithmic, and implementational levels [3]. At the computational level, attention is traditionally viewed as a mechanism to allocate limited central processing resources [4-7]. More recent studies, however, propose that attention is a mechanism to optimize task performance, even in conditions where the processing resource is not clearly constrained [8, 9]. The optimization hypothesis can explain the attention distribution in a range of well controlled learning and decision-making tasks [10, 11], but is rarely tested in complex processing tasks for which the optimal strategy is not obvious. Nevertheless, complex tasks are critical conditions to test whether the attention mechanisms abstracted from simpler tasks can truly explain real-world attention behaviors.

Reading is one of the most common and most sophisticated human behaviors [12, 13], and it is strongly regulated by attention: Since readers could only recognize a couple of words within one fixation, they have to overtly shift their attention to read a line of text [14]. Computational modeling of the reading behavior has mostly focused on normal reading of single sentences. At the computational level, it has been proposed that the eye movements are programed to, e.g., minimize the number of eye movements [15]. At the algorithmic and implementational level, models such as the E-Z reader [16] can accurately predict the eye movement trajectory with high temporal and spatial resolution. Everyday reading behavior, however, often engages reading of a multi-line passage and generally has a clear goal, e.g., information retrieval or inference generation [17]. Few models, however, have considered how the reading goal modulates reading behaviors. Here, we address this question by analyzing how readers allocate attention when reading a passage to answer a specific question in mind. The question may require, e.g., information retrieval, inference generation, or text summarization (Fig. 1). We investigate whether the task optimization hypothesis can explain the attention distribution in such goal-directed reading tasks.

Finding an optimal solution for the goal-directed reading task, however, is computationally challenging since the information related to question answering is sparsely located in a passage and their orthographic forms may not be predictable. Recent advances in DNN models, however, provide a potential tool to solve this computational problem, since DNN models equipped with attention mechanisms have approached and even surpassed mean human performance on goal-directed reading tasks [18, 19]. Attention in DNN is a mechanism to selectively extract useful information, and therefore is conceptually similar to the human attention mechanism at computational level. Furthermore, recent studies have provided strong evidence that task-optimized DNN can indeed explain the neural response properties in a range of visual and language processing tasks [20-27]. Therefore, although the DNN attention mechanism certainly deviates from the human attention mechanism in terms of its algorithms and implementation, we employ it to probe the computational-level principle underlying human attention distribution during real-world goal-directed reading.

Here, we employed DNNs to derive the optimal attention distribution for the goal-directed reading task, and tested whether such optimal distribution could explain human attention measured by eye tracking. Critically, we investigated how the attention distribution evolved along the processing hierarchy for both humans and DNNs, e.g., how text properties and the top-down task differentially modulated attention at each processing stage. Additionally, we recruited both native and non-native readers to probe how language proficiency contributed to the computational optimality of attention distribution.

**Results**
**Experiment 1: Task and Performance**

In Experiment 1, the participants ($N = 25$ for each question) first read a question and then read a passage based on which the question should be answered (Fig. 1A). After reading the passage, the participants chose from 4 options which option was the most suitable answer to the question. In total, 800 question/passage pairs were adapted from the RACE dataset [28], a collection of reading comprehension questions designed for Chinese high school students who learn English as a second language.

The questions fell into 6 types (Fig. 1BC): Three types of questions required attention to details, e.g., retrieving a fact or generate inference based on a fact, which were referred to as local questions. The other 3 types of questions concerned the general understanding of a passage, e.g., summarizing the main idea or identifying the purpose of writing, which were referred to as global questions. None of the question directly appeared in the passage, and the longest string that overlapped in the passage and question was 1.8 ± 1.5 words on average.

Participants in Experiment 1 were Chinese college or graduate students who had relatively high English proficiency. The participants correctly answered 77.94% questions on average and the accuracy was comparable across the 6 types of questions (Fig. 1B). We employed computational models to analyze what kinds of computations were required to answer the questions. The simplest heuristic model chose the option that best matched the passage orthographically (Fig. S1A). This orthographic model achieved 25.6% accuracy (Fig. 1B). Another simple heuristic model only considered word-level semantic matching between the passage and option, and achieved 27.3% accuracy (Fig. 1B). The low accuracy of the two models indicated that the reading comprehension questions could not be answered by word-level orthographic or semantic matching.

Next, we evaluated the performance of 4 context-dependent DNN models, i.e., Stanford Attentive Reader (SAR) [29], BERT [30], ALBERT [18], and RoBERTa [19], which could integrate information across words to build passage-level semantic representations. The SAR used the bi-directional recurrent neural network (RNN) to integrate contextual information (Fig. S1B) and achieved 47.6% accuracy. The other 3 models, i.e., BERT, ALBERT, and RoBERTa, were transformer-based models that were trained in 2 steps, i.e., pre-training and fine-tuning (Fig. 1D). Since the 3 models had similar structures, we averaged the performance over the 3 models (see Fig. S2 for the results of individual models). The model performance on the reading task was 37.08% and 73%, respectively, after pre-training and fine-tuning (Fig. 1B).

**Computational Models of Human Attention Distribution**

In Experiment 1, participants were allowed to read each passage for 2 minutes, but the reward they would receive was disproportional to the reading time to encourage them to develop an effective reading strategy. The results showed that the participants spent, on average, 0.7 ± 0.2 minutes reading each passage (Fig. 1C), corresponding to a reading speed of 457 ± 142 words/minute when divided by the number of words per passage. The speed was almost twice the normal reading speed for native readers [14], indicating a specialized reading strategy for the task.

Next, we employed eye tracking to quantify how the readers allocated their attention to achieve effective reading and analyze which computational models could explain the reading time on each word, i.e., the total fixation duration on each word during passage reading. In other words, we probed into what kind of computational principles could generate human-like attention distribution during goal-directed reading. A simple heuristic strategy was to attend to words that were orthographically or semantically similar to the words in the question (Fig. S1A). The predictions of the heuristic models were not highly correlated with the human word reading time (Fig. S3A, prediction accuracy around 0.2).

The DNN models analyzed here, i.e., SAR, BERT, ALBERT, and RoBERTa, all employed the attention mechanism to integrate over context to find optimal question answering strategies. Roughly speaking, the attention mechanism applied a weighted integration across all input words to generate a passage-level representation and decide whether an option was correct or not, and the weight on each word was referred to as the attention weight (see Fig. S1B and Fig. 2B for illustrations about the attention mechanisms in the SAR and transformer-based models, respectively). When the attention weights of the SAR were used to predict the human word reading time, the prediction accuracy was about 0.1 (Fig. 3A, Table S1).

In contrast to assigning a single weight on a word, the transformer-based model employed a multi-head attention mechanism: Each of the 12 layers had 12 parallel attention modules, i.e., heads. Consequently, each word had 144 attention weights (12 layers $\times$ 12 heads), which were used to model the word reading time of humans based on linear regression. Since the attention weights of 3 transformer-based models showed comparable power to predict human word reading time, we reported the prediction accuracy averaged over models (see Fig. S3A for the results of individual models). When the attention weights of pre-trained transformer-based models were used to predict the human word reading time, the prediction accuracy was around 0.5, significantly higher than the prediction accuracy of heuristic models and the SAR (Fig. 3A, Table S1). The prediction accuracy was further boosted for local but not global questions when the models were fine-tuned to perform the goal-directed reading task (Fig. 3A, Table S1). These results suggested that the human attention distribution was consistent with the attention weights in transformer-based models that were optimized to perform the same goal-directed reading task.

**Factors Influencing Human Word Reading Time**

The attention weights in transformer-based DNN models could predict the human word reading time. Nevertheless, it remained unclear whether such predictions were purely driven by basic text features that were known to modulate word reading time. Therefore, in the following, we first analyzed how basic text features modulated the word reading time during the goal-directed reading task, and then checked whether transformer-based DNNs could capture additional properties of the word reading time that could not be explained by basic text features.

Here, we further decomposed text features into visual layout features, i.e., position of a word on the screen, and word features, e.g., word length, frequency, and surprisal. Layout features were features that were mostly induced by line changes, which could be extracted without recognizing the words, while word features were finer-grained features that could only be extracted when the word or neighboring words were fixated. Linear regression analyses revealed layout features could significantly predict the word reading time (Fig. 3B, Table S2). Furthermore, the prediction accuracy was higher for global than local questions ($P = 9 \times 10^{-5}$, bootstrap, FDR corrected), suggesting a question-type-specific reading strategy. Word features could also significantly predict human reading time, even when the influence of layout features was regressed out. The predictive accuracy of the layout and word features, however, was lower than the predictive accuracy of attention weights of transformer-based models ($P = 9 \times 10^{-5}$, bootstrap, FDR corrected).

When the layout and word features were regressed out, the residual word reading time was still significantly predicted by the attention weights in transformer-based models (Fig. S3B, prediction accuracy about 0.3). This result indicated that what the transformer-based models extracted were more than basic text features. Next, we analyzed whether the transformer-based models, as well as the human word reading time, were sensitive to task-related features. To characterize the relevance of each word to the question answering task, we asked another group of participants to annotate which words contributed most to question answering. The annotated question relevance could significantly predict word reading time, even when the influences of layout and word features were regressed out (Fig. 3B, Table S2). When the question relevance was also regressed out, the residual word reading time was still significantly predicted by the attention weights in transformer-based models (Fig. S3C, P = 0.003, bootstrap, FDR corrected), but the predictive accuracy dropped to about 0.2. These results demonstrated that the DNN attention weights provided additional information about the human word reading time than the text-related and task-related features analyzed here.

Further analyses revealed two properties of the distribution of question-relevant words. First, for local questions, the question-relevant words were roughly uniformly distributed in the passage, while for global questions, the question-relevant words tended to be near the passage beginning (Fig. S4A). The eye tracking data showed that readers also spent more time reading the passage beginning for global than local questions (Fig. 3C), explaining why layout features more strongly influenced the answering of global than local questions. Second, few lines in the passage were question relevant (Fig. S4B), and the eye tracking data showed that readers spent more time reading the line with the highest question relevance (Fig. 3D), confirming the influence of question relevance on word reading time.

**Attention in Different Processing Stages for Humans and DNNs**

Next, we investigated whether humans and DNNs attended to different features in different processing stages. The early stage of human reading was indexed by the gaze duration, i.e., duration of first-pass reading of a word, and the later stage was indexed by the counts of rereading. Results showed the influence of layout features increased from early to late reading stages for global but not local questions (Fig. 4A, Table S3). Consequently, the passage-beginning-effect differed between global and local questions only for the late reading stage (Fig. S5A). The influence of word features did not strongly change between reading stages, while the influence of question relevance significantly increased from early to late reading stages (Fig. 4A, Fig. S5B). These results suggested that attention to basic text features developed early, while the influence of task mainly influenced late reading processes.

In the following, we further investigated whether transformer-based DNN attended to different features in different layers, which represented different processing stages. This analysis did not include layout features that were not available to the models. The attention weights in shallow layers were sensitive to word features in both pre-trained and fine-tuned models (Fig. 4BC). Only in the fine-tuned models, however, the attention weights in deep layers were sensitive to question relevance (see Figs. S6 & S7 for results of individual models). Therefore, the shallow and deep layers separately evolved text-based and goal-directed attention, and goal-directed attention was induced by fine-tuning on the task.

**Experiment 2: Question-Type-Specificity of the Reading Strategy**

In Experiment 1, different types of questions were presented in blocks which encouraged the participants to develop question-type-specific reading strategies. Next, we ran Experiment 2 in which questions from different types were mixed and presented in a randomized order. Since it was time consuming to measure the response to all 800 questions, we randomly selected 96 questions for Experiment 2 (16 questions per type). In Experiment 2, the reading speed was on average 298 ± 123 words/minute, lower than the speed in Experiment 1 ($P = 6 \times 10^{-4}$, bootstrap, FDR corrected), but still much faster than normal reading speed [14].

The word reading time was better predicted by fine-tuned than pre-trained transformer-based models (Fig. 5A, Table S4). For the influence of text and task-related features, compared to Experiment 1, the prediction accuracy in Experiment 2 was higher for layout and word features, but lower for question relevance (Fig. 5B, Table S5). The passage beginning effect was higher for global than local questions (Fig. 5C, 2$^{nd}$ column, $P = 2 \times 10^{-4}$, bootstrap, FDR corrected), but the difference was smaller than in Experiment 1 (Fig. 5C & Fig. S8A, $P = 2 \times 10^{-4}$, bootstrap, FDR corrected). The question relevance effect was also smaller in Experiment 2 than Experiment 1 (Fig. 5D & Fig. S8B, $P = 2 \times 10^{-4}$, bootstrap, FDR corrected). All these results indicated that the readers developed question-type-specific strategies in Experiment 1, which led to faster reading speed and stronger task modulation of word reading time.

**Experiment 3: Effect of Language Proficiency**

Experiments 1 and 2 recruited L2 readers. To investigate how language proficiency influenced task modulation of attention, we ran Experiment 3, which was the same as Experiment 2 except that the participants were native English readers. In Experiment 3, the reading speed was on average 506 ± 155 words/minute, higher than that in Experiment 2 ($P = 6 \times 10^{-4}$, bootstrap, FDR corrected). The question answering accuracy was comparable to L2 readers (Fig. 1B).

The word reading time for native readers was slightly better predicted by fine-tuned than pre-trained transformer-based models (Fig. 5A, Table S4). For the influence of text and task-related features, compared to Experiment 2, the prediction accuracy in Experiment 3 was higher for word features, but lower for layout features and question relevance (Table S5). The passage beginning effect was higher for global than local questions, but the difference was smaller than in Experiment 2 (Fig. 5C & Fig. S8A, $P = 2 \times 10^{-4}$, bootstrap, FDR corrected). The question relevance effect was also smaller for Experiment 3 than Experiment 2 (Fig. 5D & Fig. S8B, $P = 2 \times 10^{-4}$, bootstrap, FDR corrected). These results showed that the word reading time of native readers was significantly modulated by the task, but the effect was weaker than that on L2 readers.

**Experiment 4: General-Purpose Reading**

In the goal-directed reading task, participants read a passage to answer a question that they knew in advance, and the eye tracking results revealed that participants spent more time reading question-relevant words. Question-relevant words, however, were generally longer content words (Fig. S3CD) that were often associated with longer reading time even without a task [14]. Therefore, to validate the question relevance effect, we ran Experiment 4 in which the participants read the passages without knowing the question to answer. The experiment used the same 96 questions as in Experiments 2 and 3, but adopted a different experimental procedure: Participants previewed a passage before reading the question, and were allowed to read the passage again to answer the question. We analyzed the reading pattern during passage preview, which was referred to as general-purpose reading.

The participants were given 1.5 minutes to preview the passage, and the reading speed was on average 225 ± 40 words/minute, lower than that in Experiments 1-3 (P = 6 × $10^{-4}$, bootstrap, FDR corrected). Before question answering, they were given another 0.5 minutes to reread the passage, but on average they spent only 0.04 minute on rereading it. During passage preview, the word reading time was similarly predicted by the pre-trained and fine-tuned transformer-based models (Fig. 5A, Table S4). Furthermore, the word reading time was significantly predicted by layout and word features, but not question relevance (Fig. 5B, Table S4). The passage beginning effect was not significantly different between local and global questions (Fig. 5C, $4^{th}$ column, P = 0.994, bootstrap, FDR corrected), and the question relevance effect was significantly smaller than the question relevance effect in Experiments 1-3 (Fig. 5D & Fig. S8B, P = 2 × $10^{-4}$, bootstrap, FDR corrected). These results confirmed that the question relevance effects observed during goal-directed reading were indeed task dependent.

**Discussion**

Attention is a crucial mechanism to regulate information processing in the brain and it has been hypothesized that a common computational role of attention is to optimize task performance. Previous support for the hypothesis mostly comes from tasks for which the optimal strategy can be easily derived. The current study, however, considers a real-world reading task in which the participants have to actively sample a passage to answer a question that cannot be answered by simple word-level orthographic or semantic matching. In this challenging task, it is demonstrated that human attention distribution can be explained by the attention weights in transformer-based DNN models that are optimized to perform the same reading task but blind to the human eye tracking data. Furthermore, when participants scan a passage without knowing the question to answer, their attention distribution can also be explained by transformer-based DNN models that are optimized to predict a word based on the context.

Furthermore, we demonstrate that both humans and transformer-based DNN models achieve task-optimal attention distribution in multiple steps: For humans, basic text features strongly modulate the duration of the first reading of a word, while the question relevance of a word only modulates how many times the word is reread, especially for high-proficiency L2 readers compared to native readers. Similarly, for DNN models, basic text features mainly modulate the attention weights in shallow layers, while the question relevance of a word modulates the attention weights in deep layers, reflecting hierarchical control of attention to optimize task performance.

**Computational models of attention**

A large number of computational models of attention have been proposed. According to Marr's 3 levels of analysis [3], some models investigate the computational goal of attention [8, 15] and some models provide an algorithmic implementation of how different factors modulate attention [16, 31]. Computationally, it has been hypothesized that attention can be interpreted as a mechanism to optimize learning and decision making, and empirical evidence has been provided that the brain allocates attention among different information sources to optimally reduce the uncertainty of a decision [8, 9, 15]. The current study provides critical support to this hypothesis in a real-world task that engages multiple forms of attention, e.g., attention to visual layout features, attention to word features, and attention to question-relevant information. These different forms of attention, which separately modulate different eye tracking measures (Fig. 4A), jointly achieve an attention distribution that is optimal for question answering.

The transformer-based DNN models analyzed here are optimized in two steps, i.e., pre-training and fine-tuning. The results show that pre-training leads to text-based attention that can well explain general-purpose reading in Experiment 4, while the fine-tuning process leads to goal-directed attention in Experiments 1-3 (Fig. 4B & Fig. 5A). Pre-training is also achieved through task optimization, and the pre-training task used in all the three models analyzed here is to predict a word based on the context. The purpose of the word prediction task is to let models learn the general statistical regularity in a language based on large corpora, and this process is crucial for model performance on downstream tasks [18, 19, 30]. Previous eye-tracking studies have suggested that the predictability of words, i.e., surprisal, can modulate reading time [32], and neuroscientific studies have also indicated that the cortical responses to language converge with the representations in pre-trained DNN models [22, 23]. The results here further demonstrate that the DNN optimized for the word prediction task can evolve attention properties consistent with the human reading process.

A separate class of models investigates which factors shape human attention distribution. A large number of models are proposed to predict bottom-up visual saliency [33, 34], and recently DNN models are also employed to model top-down visual attention. It is shown that, through either implicit [35, 36] or explicit training [37], DNNs can predict which parts of a picture relate to a verbal phrase, a task similar to goal-directed visual search [38]. The current study distinguishes from these studies in that the DNN model is not trained to predict human attention. Instead, the DNN models naturally generate human-like attention distribution when trained to perform the same task that humans perform, suggesting that task optimization is a potential cause for human attention distribution during reading.

**Models for human reading and human attention to question-relevant information**

How human readers allocate attention during reading is an extensively studied topic, mostly based on studies that instruct readers to read a sentence in a normal manner, not aimed to extract a specific kind of information [39]. Previous eye tracking studies have shown that the readers fixate longer upon, e.g., longer words, words of lower-frequency, words that are less predictable based on the context, and words at the beginning of a line [14]. A number of models, e.g., the E-Z reader [16] and SWIFT [40], have been proposed to predict the eye movements during reading based on basic oculomotor properties or lexical processing [16]. Some models also view reading as an optimization process that minimizes the time or the number of saccades required to read a sentence [15, 41]. These models can generate fine-grained predictions, e.g., which letter in a word will be fixated first, for the reading of simple sentences, but have only been occasionally tested for complex sentences or multi-line texts [42] or to characterize different reading tasks, e.g., z-string reading and visual searching [43]. A recent model has also considered the specific reading goal of the participants [44], and can explain the word reading time when the readers read a passage to answer a relatively simple question that can be answered using a word-matching strategy [45]. Future studies can potentially integrate classic eye movement models with DNNs to explain the dynamic eye movement trajectory, possibly with a letter-based spatial resolution.

When human readers read a passage with a particular goal or perspective, previous studies have revealed inconsistent results about whether the readers spent more time reading task-relevant sentences [46-48]. To explain the inconsistent results, it has been proposed that the question relevance effect weakens for readers with a higher working memory and when readers read a familiar topic [49]. Similarly, here, we demonstrate that non-native readers indeed spend more time reading question-relevant information than native readers do (Fig. 5D & Fig. S8B). Therefore, it is possible that when readers are more skilled and when the passage is relatively easy to read, their processing is so efficient so that they do not need extra time to encode task-relevant information.

**DNN attention to question-relevant information**

A number of studies have investigated whether the DNN attention weights are interpretable, but the conclusions are mixed: Some studies find that the DNN attention weights are positively correlated with the importance of each word [50, 51], while other studies fail to find such correlation [52, 53]. The inconsistent results are potentially caused by the lack of gold standard to evaluate the contribution of each word to a task. A few recent studies have used the human word reading time as the criterion to quantify word importance, but these studies do not reach consistent conclusions either. Some studies find that the attention weights in the last layer of transformer-based DNN models better correlates with human word reading time than basic word frequency measures [54], and integrating human word reading time into DNN can slightly improve task performance [55]. Other studies, however, find no meaningful correlation between the attention weights in transformer-based DNNs and human word reading time [56].

The current results provide a potential explanation for the discrepancy in the literature: The last layer of transformer-based DNNs is tuned to task relevant information (Fig. 4B), but the influence of task relevance on word reading time is rather weak for native readers (Fig. 5B). Consequently, the correlation between the last-layer DNN attention weights and human reading time may not be robust. The current results demonstrate that the reading time of both native and non-native readers are reliably modulated by basic text features, which can be modeled by the attention weights in shallower DNN layers.

Finally, the current study demonstrates that transformer-based DNN models can automatically generate human-like attention, in the absence of any prior knowledge about the properties of the human reading process. Simpler models that fail to explain human performance also fail to predict human attention distribution. It remains possible, however, different models can solve the same computational problem using distinct algorithms, and only some algorithms generate human-like attention distribution. In other words, human-like attention distribution may not be a unique solution to optimize the goal-directed reading task. Sharing similar attention distribution with humans, however, provides a way to interpret the attention weights in computational models. From this perspective, the dataset and methods developed here provides an effective probe to test the biological plausibility of NLP models that can be easily applied to test whether a model evolves human-like attention distribution.

## Materials and Methods

### Participants

Totally, 162 participants took part in this study (19-30 years old, mean age, 22.5 years; 84 female). All participants had normal or corrected-to-normal vision. Experiment 1 had 102 participants. Experiments 2-4 had 20 participants. No participant took part in more than one experiment. Additional 17 participants were recruited but failed to pass the calibration process for eye tracking and therefore did not participate in the reading experiments.

In Experiments 1, 2 and 4, participants were native Chinese readers. They were college students or graduate students from Zhejiang University, and were thus above the level required to answer high-school-level reading comprehension questions. English proficiency levels were further guaranteed by the following criterion for screening participants: a minimum score of 6 on IELTS, 80 on TOEFL, or 425 on CET6[1]. In Experiment 3, participants were native English readers. The experimental procedures were approved by the Research Ethics Committee of the College of Medicine, Zhejiang University (2019–047). The participants provided written consent and were paid.

**Experimental materials**

The reading materials were selected and adapted from the large-scale RACE dataset, a collection of reading comprehension questions in English exams for middle and high schools in China [28]. We selected 800 high-school level questions from the test set of RACE and each question was associated with a distinct passage (117 to 456 words per passage). All questions were multiple-choice questions with 4 alternatives including only one correct option among them. The questions fell into 6 types, i.e., Cause ($N = 200$), Fact ($N = 200$), Inference ($N = 120$), Theme ($N = 100$), Title ($N = 100$), and Purpose ($N = 80$). The Cause, Fact, and Inference questions concerned the location, extraction, and comprehension of specific information from a passage, and were referred to as local questions. Questions of Theme, Title, and Purpose tested the understanding of a passage as a whole, and were referred to as global questions.

In a separate online experiment, we acquired annotations about the relevance of each word to the question answering task. For each passage, a participant was allowed to annotate up to 5 key words that were considered relevant to answering the corresponding question. Each passage was annotated by *N* participants (*N* ≥ 26), producing *N* versions of annotated key words. Each version of annotation was then validated by a separate participant. In the validation procedure, the participant was required to answer the question solely based on the key words of a specific annotation version; if the person could not derive the correct answer, this version of annotation was discarded. The percentage of questions correctly answered in the validation procedure was 75.9% and 67.6%, for local and global questions respectively. If *M* versions of annotation passed the validation procedure and a word was annotated in *K* versions, the question relevance of the word was *K/M*. More details about the question types and the annotation procedures could be found in the reference [57].

---

[1] The National College English Test (CET) is a national English test system developed to examine the English proficiency of college students in China. The CET includes tests of two levels: a lower level test CET4 and a higher level test CET6.

**Experimental procedures**

**Experiment 1:** Experiment 1 included all 800 passages, and different question types were separately tested in different sessions, hence 6 sessions in total. Each session included 25 participants and one participant could participate in multiple sessions. Before each session, participants were familiarized with 5 questions that were not used in the formal session. During the formal session, questions were presented in a randomized order. Considering the quantities of questions, for Cause and Fact questions, the session was carried out in 3 separate days (one third questions on each day), and for other question types, the session was carried out in 2 separate days (fifty percent of questions on each day).

The experiment procedure in Experiment 1 was illustrated in Fig. 1A. In each trial, participants first read a question, pressed the space bar to read the corresponding passage, pressed the space bar again to read the question coupled with 4 options, and chose the correct answer. The time limit for passage reading was 120 s. To encourage the participants to read as quickly as possible, the bonus they received for a specific question would decrease linearly from 1.5 to 0.5 RMB over time. They did not receive any bonus for the question, however, if they gave a wrong answer. Furthermore, before answering the comprehension question, the participants reported whether they were confident about that they could correctly answer the question (yes or no). Participants selected yes for 90.47% of questions (89.62% and 92.04% for local and global questions, respectively). After answering the question, they also rated their confidence about their answer on the scale of 1-4 (low to high). The mean confidence rating was 3.25 (3.28 and 3.18 for local and global question, respectively), suggesting that the participants were confident about their answers.

**Experiments 2 and 3:** Experiments 2 and 3 included 96 reading passages and questions that were randomly selected from the questions used in Experiment 1 and included 16 questions for each question type. The 6 types of questions were mixed and presented in a randomized order. The trial structure, as well as the familiarization procedure, in Experiments 2 and 3 was identical to that in Experiment 1. Experiments 2 and 3 were identical except that Experiment 2 recruited high-proficiency L2 readers while Experiment 3 recruited native English readers.

**Experiment 4:** Experiment 4 included the 96 questions presented in Experiments 2 and 3, which were presented in a randomized order. The trial structure in Experiment 4 is similar to that in Experiments 1-3, except that a 90-s passage preview stage was introduced at the beginning of each trial. During passage preview, participants had no prior information of the relevant question. The participants could press the space bar to terminate the preview and to read a question. Then, participants read the passage again with a time limit of 30 s, before proceeding to answer the question. The payment method was similar to Experiment 2, and the bonus was calculated based on the duration of second-pass passage reading.

**Stimulus presentation and eye tracking**

The text was presented using the bold Courier New font, and each letter occupied $14 \times 27$ pixels. We set the maximum number of letters on each line to 120 and used double space. We separated paragraphs by indenting the first line of each new paragraph. Participants sat about 880 mm from a monitor, at which each letter horizontally subtended approximately 0.25 degrees of visual angle.

Eye tracking data were recorded from the left eye with 500-Hz sampling rate (Eyelink Portable Duo, SR Research). The experiment stimuli were presented on a 24-inch monitor ($1920 \times 1080$ resolution; 60 Hz refresh rate) and administered using MATLAB Psychtoolbox [58]. Each experiment started with a 13-point calibration and validation of eye tracker, and the validation error was required to be below 0.5 degrees of visual angle. Furthermore, before each trial, a 1-point validation was applied, and if the calibration error was higher than 0.5 degrees of visual angle, a recalibration was carried out. Head movements were minimized using a chin and forehead rest.

**Word-level reading comprehension models**

The orthographic and semantic models probed whether the reading comprehension questions could be answered based on word-level orthographic or semantic information. Both models calculated the similarity between each content word in the passage and each content word in an option, and averaged the word-by-word similarity across all words in the passage and all words in the option (Fig. S1A). The option with the highest mean similarity value was chosen as the answer. For the orthographic model, similarity was quantified using the edit distance [59]. For the semantic model, similarity was quantified by the correlation between vectorial representations of word meaning, i.e., the glove model [60]. Performance of the models remained similar if the answer was chosen based on the maximal word-by-word similarity, instead of the mean similarity.

**RNN-based reading comprehension models**

The SAR was a classical RNN-based model for the reading comprehension task [29]. In contrast to the word-level models, the SAR was context sensitive and employed bi-directional RNNs to integrate information across words (Fig. S1B). Independent bi-directional RNNs were employed to build a vectorial representation for the question and each option. An additional bi-directional RNN was applied to construct a vectorial representation for each word in the passage, and a passage representation was built by a weighted sum of the representations of individual words in the passage. The weight on each word, i.e., the attention weight, captured the similarity between the representation of the word and the question representation using a bilinear function. Finally, based on the passage representation and each option representation, a bilinear dot layer calculated the possibility that the option was the correct answer.

**Transformer-based reading comprehension models**

We tested 3 popular transformer-based DNN models, i.e., BERT [30], ALBERT [18], and RoBERTa [19], which were all reported to reach high performance on the reading comprehension task. ALBERT and RoBERTa were both adapted from BERT, and had the same basic structure. RoBERTa differed from BERT in its pre-training procedure [19] while ALBERT applied factorized embedding parameterization and cross-layer parameter sharing to reduce memory consumption [18]. Following previous studies [18, 19], each option was independently processed. For the $i^{th}$ option ($i$ = 1, 2, 3, or 4), the question and the option were concatenated to form an integrated option. As shown in the left panel of Fig. 2B, for the $i^{th}$ option, the input to models was the following sequence:

$$CLS_i, P_1, P_2, ..., P_N, S_{i,1}, O_{i,1}, O_{i,2}, ..., O_{i,M}, S_{i,2},$$

where $CLS_i$, $S_{i,1}$, and $S_{i,2}$ denoted special tokens separating different components of the input. $P_1, P_2, \ldots, P_N$ denoted all the $N$ words of a passage, and $O_{i,1}, O_{i,2}, \ldots, O_{i,M}$ denoted all the $M$ words in the $i^{th}$ integrated option. Each of the token was represented by a vector. The vectorial representation was updated in each layer, and in the following the output of the $l^{th}$ layer was denoted as a superscript, e.g., $CLS_i^l$. Following previous studies [18, 19], we calculated a score for each option, which indicated the possibility that the option was the correct answer. The score was calculated by first applying a linear transform to the final representation of the CLS token, i.e.,

$$s_i = \Phi CLS_i^{12},$$

where $CLS_i^{12}$ was the final output representation of CLS and $\Phi$ was a vector learned from data. The score was independently calculated for each option and then normalized using the following equation:

$$score_i = \frac{exp(s_i)}{\sum_{i=1}^{4} exp(s_i)}.$$

The answer to a question was determined as the option with the highest score, and all the models were trained to maximize the logarithmic score of the correct option. The transformer-based models were trained in two steps (Fig. 1D). The pre-training process aimed to learn general statistical regularities in a language based on large corpora, while the fine-tuning process trained models to perform the reading comprehension task. All models were implemented based on HuggingFace [61] and all hyperparameters for fine-tuning were adopted from previous studies [18, 19, 62, 63] (see Table S6).

**Attention in transformer-based models**

The transformer-based models we applied had 12 layers, and each layer had 12 parallel attention heads. Each attention head calculated an attention weight between any pair of inputs, including words and special tokens. The vectorial representation of each input was then updated by the weighted sum of the vectorial representations of all inputs [64]. Since only the CLS token was directly related to question answering, here we restrained the analysis to the attention weights that were used to calculate the vectorial representation of CLS (Fig. 2B, right panel). In the $h^{th}$ head, the vectorial representation of CLS was computed using the following equations. For the sake of clarity, we did not distinguish the input words and special tokens and simply denoted them as $X_i$.

$$CLS^h = \sum_{i=1}^{N+M+3} \alpha_i V_i = \alpha_{CLS} V_{CLS} + \sum_{n=1}^{N} \alpha_{Pn} V_{Pn} + \alpha_{S1} V_{S1} + \sum_{m=1}^{M} \alpha_{Om} V_{Om} + \alpha_{S2} V_{S2},$$

$$\alpha_i = \frac{exp(Q_{CLS} K_i^T)}{\sum_{i=1}^{N+M+3} exp(Q_{CLS} K_i^T)},$$

$$V_i = X_i W^V + b^V, \quad K_i = X_i W^K + b^K, \quad Q_{CLS} = X_{CLS} W^Q + b^Q,$$

where $W^V$, $W^Q$, $W^K$, $b^V$, $b^Q$, and $b^K$ were parameters to learn from the data, and $\alpha_i$ was the attention weight between CLS and $X_i$. The attention weight between CLS and the $n^{th}$ word in the passage, i.e., $α_{Pn}$, was compared to human attention. Here, we only considered the attention weights associated with the correct option. Additionally, DNNs used byte-pair tokenization which split some words into multiple tokens. We converted the token-level attention weights to word-level attention weights by summing the attention weights over tokens within a word [54, 65].

**Eye tracking measures**

We analyzed eye movements during passage reading in Experiments 1-3, and the passage preview in Experiment 4. For each word, the total fixation time, gaze duration, and run counts was extracted using the SR Research Data Viewer software. The total fixation time of a word was referred to as the word reading time. The gaze duration was the how long a word was fixated before the gaze moved to other words, reflected first-pass processing of a word. To characterize late processing of a word, we further calculated the counts of rereading, which were defined as the run counts minus 1. Words that were not reread were excluded from the analysis of counts of rereading. Each of the eye tracking measure was averaged across all participants who correctly answered the question.

**Regression models**

We employed linear regression to analyze how well each model, as well as each set of text/task-related features, could explain human attention measured by eye tracking. In all regression analyses, each regressor and the eye-tracking measure were normalized within each passage by taking the z-score. The prediction accuracy, i.e., the correlation between the predicted eye-tracking measure and the actual eye-tracking measure was calculated based on five-fold cross-validation.

*Regressors:* For the SAR, each word had one attention weight, which was used as the regressor. For transformer-based models, since each model contained 12 layers and each layer contained 12 attention heads, all together there were 144 regressors. Text features included layout features and word features. The layout features concerned the visual position of text, including the coordinate of the left most pixel of a word, ordinal position of a word in a paragraph, ordinal line number of a word in a paragraph, and ordinal line number of a word in a passage. The word features included word length, logarithmic word frequency estimated based on the British National Corpus [66], and word surprisal estimated from a trigram language model with Kneser-Ney smoothing trained on British National Corpus using SRILM [67]. The task-related feature referred to the question relevance annotated by another group of participants (see Experimental materials for details).

Additionally, we also applied linear regression to probe how DNN attention was affected by text features and question relevance. Since information of lines and paragraphs were not available to DNNs, the layout features only included the ordinal position of a word in a sentence, ordinal position of a word in a passage, and ordinal sentence number of a word in this analysis.

**Statistical tests**

In the regression analysis, we employed a one-sided permutation test to test whether a set of features could statistically significantly predict an eye tracking measure. Five hundred chance-level prediction accuracy was calculated by predicting the eye tracking measure shuffled across all words within a passage: The eye tracking measure to predict was shuffled but the features were not. The procedure was repeated 500 times, creating 500 chance-level prediction accuracy. If the actual correlation was greater than *N* out of the 500 chance-level correlation, the significance level was (*N* + 1)/501.

When comparing the responses to local and global questions, the 3 types of local/global questions were pooled. The comparison between local and global questions, as well as the comparison between experiments, was based on bias-corrected and accelerated bootstrap [68]. For example, to test whether the prediction accuracy differed between the 2 types of questions, all global questions were resampled with replacement 50000 times and each time the prediction accuracy was calculated based on the resampled questions, resulting in 50000 resampled prediction accuracy. If the prediction accuracy for local questions was greater (or smaller) than *N* out of the 50000 resampled accuracy for global questions, the significance level of their difference was 2(*N* + 1)/50001. When multiple comparisons were performed, the p-value was further adjusted using the false discovery rate (FDR) correction.

**Acknowledgements**

We thank David Poeppel, Yunyi Pan, and Erik D. Reichle for valuable comments on earlier versions of this manuscript; Jonathan Simon, Bingjiang Lyu, and members of the Ding lab for thoughtful discussions and feedback; Qian Chu, Yuhan Lu, Anqi Dai, Zhonghua Tang, and Yan Chen for assistance with experiments. Work supported by National Key Research and Development Program of China 2021ZD0204105，National Natural Science Foundation of China 32222035, and Major Scientific Research Project of Zhejiang Lab 2019KB0AC02.

**Author Contributions**

Nai Ding acquired the funding, conceived and coordinated the project, analyzed data, and wrote the manuscript. Xing Tian coordinated the project and revised the manuscript. Jiajie Zou implemented the experiments and models, analyzed data, and wrote the manuscript. Yuran Zhang and Jialu Li implemented the experiments.

**Competing Interest Statement**

The authors declare no competing interests.

**Data Availability**

All eye tracking data will be available before publication at

https://github.com/jiajiezou/TOA.

**References**

1. Posner MI, Petersen SE. The attention system of the human brain. Annu Rev Neurosci. 1990;13(1):25-42.

2. Treisman AM, Gelade G. A feature-integration theory of attention. Cogn Psychol. 1980;12(1):97-136.

3. Marr D. Vision: A computational investigation into the human representation and processing of visual information, henry holt and co. Inc, New York, NY. 1982;2(4.2).

4. Kahneman D. Attention and effort: Citeseer; 1973.

5. Franconeri SL, Alvarez GA, Cavanagh P. Flexible cognitive resources: competitive content maps for attention and memory. Trends in Cognitive Sciences. 2013;17(3):134-41. doi: https://doi.org/10.1016/j.tics.2013.01.010.

6. Lennie P. The Cost of Cortical Computation. Current Biology. 2003;13(6):493-7. doi: https://doi.org/10.1016/S0960-9822(03)00135-0.

7. Carrasco M. Visual attention: The past 25 years. Vision research. 2011;51(13):1484-525. doi: https://doi.org/10.1016/j.visres.2011.04.012.

8. Dayan P, Kakade S, Montague PR. Learning and selective attention. Nat Neurosci. 2000;3(11):1218-23.

9. Gottlieb J, Hayhoe M, Hikosaka O, Rangel A. Attention, reward, and information seeking. J Neurosci. 2014;34(46):15497-504. Epub 2014/11/14. PubMed PMID: 25392517; PubMed Central PMCID: PMCPMC4228145.

10. Najemnik J, Geisler WS. Optimal eye movement strategies in visual search. Nature. 2005;434(7031):387-91.

11. Navalpakkam V, Koch C, Rangel A, Perona P. Optimal reward harvesting in complex perceptual environments. Proc Natl Acad Sci USA. 2010;107(11):5232-7.

12. Li X, Huang L, Yao P, Hyönä J. Universal and specific reading mechanisms across different writing systems. Nat Rev Psychol. 2022;1:133-44.

13. Gagl B, Gregorova K, Golch J, Hawelka S, Sassenhagen J, Tavano A, et al. Eye movements during text reading align with the rate of speech production.

Nat Hum Behav. 2021;6:429-42. doi: 10.1038/s41562-021-01215-4.

14. Rayner K. Eye movements in reading and information processing: 20 years of research. Psychol Bull. 1998;124(3):372-422.

15. Legge GE, Hooven TA, Klitz TS, Mansfield JS, Tjan BS. Mr. Chips 2002: New insights from an ideal-observer model of reading. Vision Res. 2002;42(18):2219-34.

16. Reichle ED, Rayner K, Pollatsek A. The EZ Reader model of eye-movement control in reading: Comparisons to other models. Behav Brain Sci. 2003;26(4):445-76.

17. White S, Chen J, Forsyth B. Reading-related literacy activities of American adults: Time spent, task types, and cognitive skills used. J Lit Res. 2010;42(3):276-307.

18. Lan Z, Chen M, Goodman S, Gimpel K, Sharma P, Soricut R. Albert: A lite bert for self-supervised learning of language representations. International Conference on Learning Representations; 2019: ICLR; 2020.

19. Liu Y, Ott M, Goyal N, Du J, Joshi M, Chen D, et al. Roberta: A robustly optimized bert pretraining approach. arXiv. 2019. doi: arXiv:1907.11692.

20. Yamins DL, Hong H, Cadieu CF, Solomon EA, Seibert D, DiCarlo JJ. Performance-optimized hierarchical models predict neural responses in higher visual cortex. Proc Natl Acad Sci USA. 2014;111(23):8619-24.

21. Kell AJ, Yamins DL, Shook EN, Norman-Haignere SV, McDermott JH. A task-optimized neural network replicates human auditory behavior, predicts brain responses, and reveals a cortical processing hierarchy. Neuron. 2018;98(3):630-44. e16.

22. Goldstein A, Zada Z, Buchnik E, Schain M, Price A, Aubrey B, et al. Shared computational principles for language processing in humans and deep language models. Nat Neurosci. 2022;25(3):369-80.

23. Schrimpf M, Blank IA, Tuckute G, Kauf C, Hosseini EA, Kanwisher N, et al. The neural architecture of language: Integrative modeling converges on predictive processing. Proc Natl Acad Sci USA. 2021;118(45):e2105646118.

doi: 10.1073/pnas.2105646118.

24. Hasson U, Nastase SA, Goldstein A. Direct Fit to Nature: An Evolutionary Perspective on Biological and Artificial Neural Networks. Neuron. 2020;105(3):416-34. doi: 10.1016/j.neuron.2019.12.002. PubMed PMID: 32027833; PubMed Central PMCID: PMC7096172.

25. Donhauser PW, Baillet S. Two distinct neural timescales for predictive speech processing. Neuron. 2020;105(2):385-93. e9.

26. Rabovsky M, Hansen SS, McClelland JL. Modelling the N400 brain potential as change in a probabilistic representation of meaning. Nature Human Behaviour. 2018;2(9):693-705.

27. Heilbron M, Armeni K, Schoffelen J-M, Hagoort P, de Lange FP. A hierarchy of linguistic predictions during natural language comprehension. Proc Natl Acad Sci USA. 2022;119(32):e2201968119. doi: doi:10.1073/pnas.2201968119.

28. Lai G, Xie Q, Liu H, Yang Y, Hovy E. Race: Large-scale reading comprehension dataset from examinations. 2017 Conference on Empirical Methods in Natural Language Processing; 2017: ACL; 2017.

29. Chen D, Bolton J, Manning CD. A thorough examination of the cnn/daily mail reading comprehension task. 54th annual meeting of the association for computational linguistics; 2016: ACL; 2016.

30. Devlin J, Chang M-W, Lee K, Toutanova K. Bert: Pre-training of deep bidirectional transformers for language understanding. 2019 Conference of the North American Chapter of the Association for Computational Linguistics; 2019: ACL; 2019.

31. Itti L, Koch C, Niebur E. A model of saliency-based visual attention for rapid scene analysis. IEEE Trans Pattern Anal Mach Intell. 1998;20(11):1254-9.

32. Hale J. Information‐theoretical complexity metrics. Lang Linguist Compass. 2016;10(9):397-412.

33. Tatler BW, Hayhoe MM, Land MF, Ballard DH. Eye guidance in natural vision: reinterpreting salience. J Vis. 2011;11(5):5-25. Epub 2011/05/31. doi:

10.1167/11.5.5. PubMed PMID: 21622729; PubMed Central PMCID: PMCPMC3134223.

34. Borji A, Sihite DN, Itti L. Quantitative analysis of human-model agreement in visual saliency modeling: a comparative study. IEEE Trans Image Process. 2013;22(1):55-69. Epub 2012/08/08. PubMed PMID: 22868572.

35. Anderson P, He X, Buehler C, Teney D, Johnson M, Gould S, et al. Bottom-up and top-down attention for image captioning and visual question answering. The IEEE Conference on Computer Vision and Pattern Recognition; 2018: CVPR; 2018.

36. Xu K, Ba J, Kiros R, Cho K, Courville A, Salakhudinov R, et al. Show, attend and tell: Neural image caption generation with visual attention. 32nd International Conference on Machine Learning; 2015: PMLR; 2015.

37. Das A, Agrawal H, Zitnick L, Parikh D, Batra D. Human attention in visual question answering: Do humans and deep networks look at the same regions? Comput Vis Image Underst. 2017;163:90-100.

38. Wolfe JM, Horowitz TS. Five factors that guide attention in visual search. Nat Hum Behav. 2017;1(3):1-8.

39. Clifton C, Ferreira F, Henderson JM, Inhoff AW, Liversedge SP, Reichle ED, et al. Eye movements in reading and information processing: Keith Rayner's 40 year legacy. J Mem Lang. 2016;86:1-19. doi: 10.1016/j.jml.2015.07.004.

40. Engbert R, Nuthmann A, Richter EM, Kliegl R. SWIFT: A dynamical model of saccade generation during reading. Psychol Rev. 2005;112(4):777-813.

41. Liu Y, Reichle E. The emergence of adaptive eye movements in reading. Cogsci. 2010;32(32).

42. Mancheva L, Reichle ED, Lemaire B, Valdois S, Ecalle J, Guérin-Dugué A. An analysis of reading skill development using EZ Reader. J Cogn Psychol. 2015;27(5):657-76.

43. Reichle ED, Pollatsek A, Rayner K. Using EZ Reader to simulate eye movements in nonreading tasks: A unified framework for understanding the eye–mind link. Psychol Rev. 2012;119(1):155.

44. Hahn M, Keller F. Modeling task effects in human reading with neural network-based attention. Cognition. 2023;230:105289. doi: https://doi.org/10.1016/j.cognition.2022.105289.

45. Hermann KM, Kocisky T, Grefenstette E, Espeholt L, Kay W, Suleyman M, et al. Teaching machines to read and comprehend. Advances in Neural Information Processing Systems; 2015.

46. Yeari M, van den Broek P, Oudega M. Processing and memory of central versus peripheral information as a function of reading goals: Evidence from eye-movements. Read Writ. 2015;28(8):1071-97.

47. Grabe M. Reader imposed structure and prose retention. Contemporary Educational Psychology. 1979;4(2):162-71.

48. Kaakinen JK, Hyönä J, Keenan JM. Perspective effects on online text processing. Discourse processes. 2002;33(2):159-73.

49. Kaakinen JK, Hyönä J, Keenan JM. How prior knowledge, WMC, and relevance of information affect eye fixations in expository text. J Exp Psychol. 2003;29(3):447-57.

50. Yang Z, Yang D, Dyer C, He X, Smola A, Hovy E. Hierarchical attention networks for document classification. 2016 Conference of the North American Chapter of the Association for Computational Linguistics: Human Language Technologies; 2016: ACL; 2016.

51. Lin Z, Feng M, Santos CNd, Yu M, Xiang B, Zhou B, et al. A structured self-attentive sentence embedding. International Conference on Learning Representations; 2017: ICLR; 2017.

52. Serrano S, Smith NA. Is Attention Interpretable? 57th Annual Meeting of the Association for Computational Linguistics; 2019: Association for Computational Linguistics; 2019.

53. Jain S, Wallace BC. Attention is not Explanation. 2019 Conference of the North American Chapter of the Association for Computational Linguistics; 2019 jun; Minneapolis, Minnesota: Association for Computational Linguistics.

54. Bolotova V, Blinov V, Zheng Y, Croft WB, Scholer F, Sanderson M. Do

People and Neural Nets Pay Attention to the Same Words: Studying Eye-tracking Data for Non-factoid QA Evaluation. 29th ACM International Conference on Information & Knowledge Management; 2020: ACM; 2020.

55. Malmaud J, Levy R, Berzak Y. Bridging Information-Seeking Human Gaze and Machine Reading Comprehension. arXiv. 2020.

56. Sood E, Tannert S, Frassinelli D, Bulling A, Vu NT. Interpreting attention models with human visual attention in machine reading comprehension. 24th Conference on Computational Natural Language Learning; 2020: Association for Computational Linguistics.

57. Zou J, Zhang Y, Jin P, Luo C, Pan X, Ding N. PALRACE: Reading Comprehension Dataset with Human Data and Labeled Rationales2021. Available from: https://arxiv.org/abs/2106.12373.

58. Brainard DH. The psychophysics toolbox. Spat Vis. 1997;10(4):433-6.

59. Levenshtein VI. Binary codes capable of correcting deletions, insertions, and reversals. Soviet physics doklady; 1966: Soviet Union.

60. Pennington J, Socher R, Manning C. Glove: Global vectors for word representation. 2014 Conference on Empirical Methods in Natural Language Processing; 2014.

61. Wolf T, Debut L, Sanh V, Chaumond J, Delangue C, Moi A, et al. HuggingFace's Transformers: State-of-the-art natural language processing. 2020 Conference on Empirical Methods in Natural Language Processing: System Demonstrations; 2019: Association for Computational Linguistics; 2020.

62. Zhang S, Zhao H, Wu Y, Zhang Z, Zhou X, Zhou X. DCMN+: Dual co-matching network for multi-choice reading comprehension. AAAI conference on artificial intelligence; 2019: AAAI 2020.

63. Ran Q, Li P, Hu W, Zhou J. Option comparison network for multiple-choice reading comprehension. arXiv. 2019. doi: arXiv:1903.03033.

64. Vaswani A, Shazeer N, Parmar N, Uszkoreit J, Jones L, Gomez AN, et al. Attention is all you need. Advances in Neural Information Processing

Systems; 2017: Curran Associates; 2017.

65. Clark K, Khandelwal U, Levy O, Manning CD. What does BERT look at? An analysis of BERT's attention. 2019 ACL workshop blackboxNLP: Analyzing and interpreting neural networks for NLP; 2019: ACL; 2019.

66. Burnard L. The British national corpus, version 3 (BNC XML Edition). 2007. Available from: http://www.natcorp.ox.ac.uk/.

67. Stolcke A. SRILM-an extensible language modeling toolkit. Seventh International Conference on Spoken Language Processing; 2002.

68. Efron B, Tibshirani RJ. An introduction to the bootstrap: CRC press; 1994.

# Figures

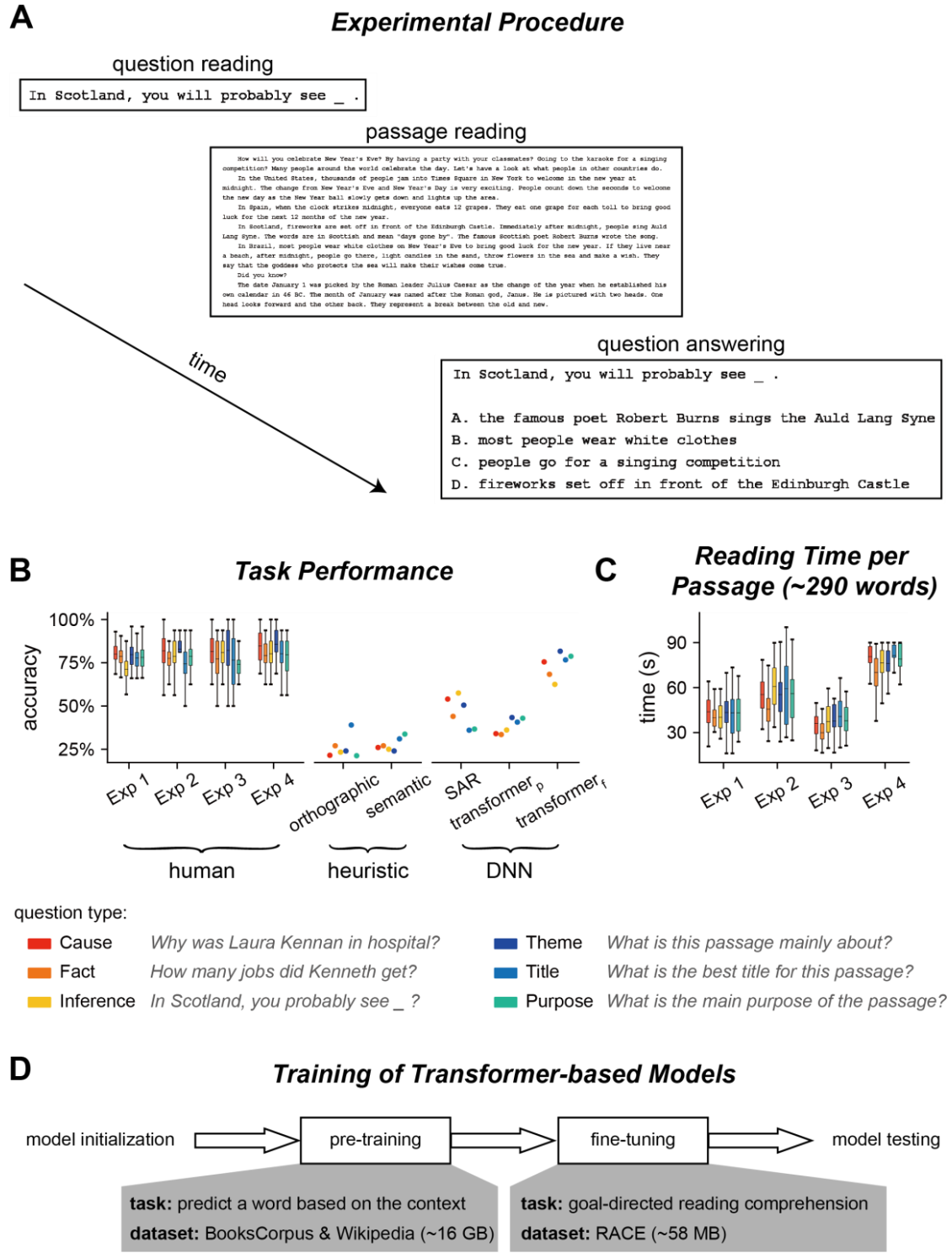

**Fig. 1. Experiment and performance.** (**A**) Experimental procedure for Experiments 1-3. In each trial, participants saw a question before reading a passage. After reading the passage, they chose the answer to the question from 4 options. (**B**) Accuracy of question answering for humans and computational models. Transformer$_p$ and

transformer$_f$ separately denote pre-trained and fine-tuned transformer-based language models. The question type is color coded and an example question is shown for each type. (**C**) Time spent on reading each passage. The box plot shows the mean (horizontal lines inside the box), 25$^{th}$ and 75$^{th}$ percentiles (box boundaries), and 25$^{th}$/75$^{th}$ percentiles ± 1.5 × interquartile range (whiskers). (**D**) Illustration of the training process for transformer-based models. The pre-training process aims to learn general statistical regularities in a language based on large corpora, while the fine-tuning process trains models to perform the reading comprehension task.

## A  *Examples of Human Attention Distribution and Model Predictions*

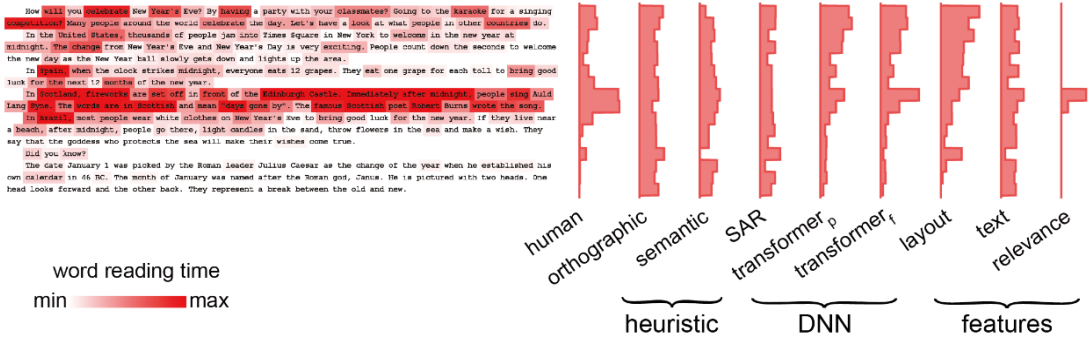

local question ("In Scotland, you will probably see __ .")

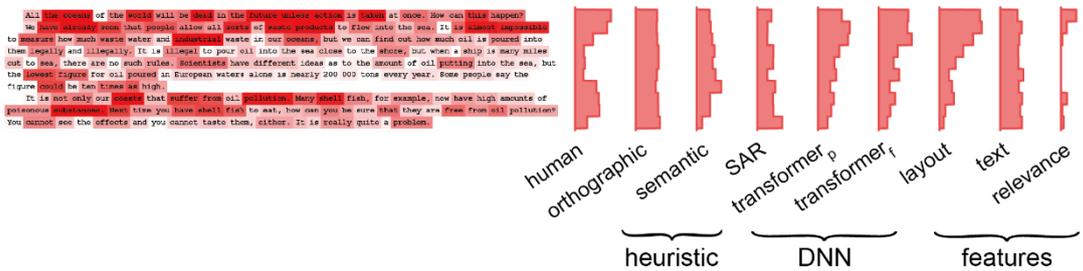

global question ("What is the passage mainly about?")

## B  *Transformer-based DNN Models*

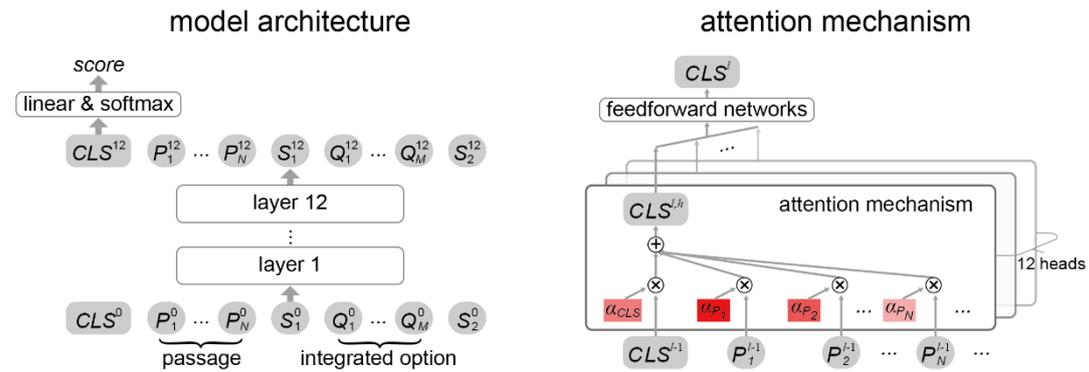

**Fig. 2. Human attention distribution and computational models.** (**A**) Examples of human attention distribution, quantified by the word reading time. The histograms on the right showed the mean reading time on each line, for both human data and model predictions. (B) The general architecture of the 12-layer transformer-based models. The model input consists of all words in the passage and an integrated option. Output of the model relies on the node $CLS^{12}$, which is used to calculate a score reflecting how likely an option is the correct answer. The CLS node is a weighted sum of the vectorial representations of all words and tokens, and the attention weight for each word in the passage, i.e., $\alpha$, is the DNN attention analyzed in this study.

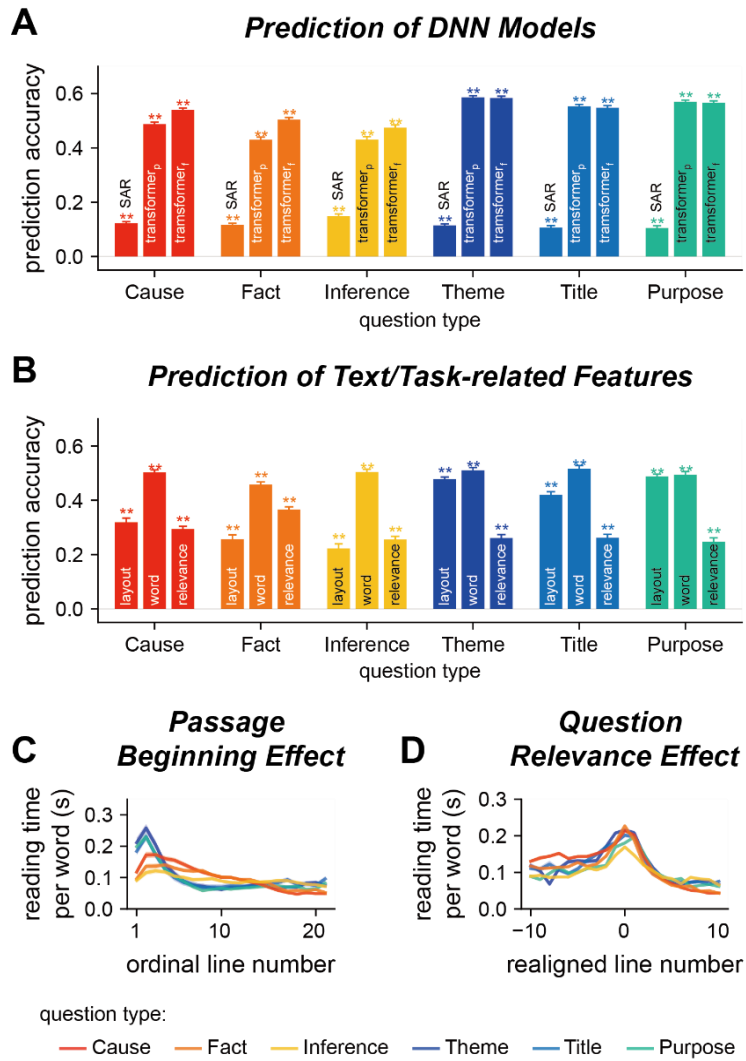

**Fig. 3. Model word reading time in Experiment 1.** (**AB**) Predict the word reading time based on the attention weights of DNN models, text features, or question relevance. The predictive accuracy is the correlation coefficient between the predicted word reading time and the actual word reading time. Prediction accuracy significantly higher than chance is denoted by stars on the top of each bar. **$P < 0.01$. (**C**) Relationship between the word reading time and line index. The word reading time is longer near the beginning of a passage and the effect is stronger for global questions than local questions. (**D**) Relationship between the word reading time and question relevance. Line 0 refers to the line with the highest question relevance. The word reading time is higher for the question-relevant line. Color indicates the question type. The shade area indicates one standard error of the mean (SEM) across participants.

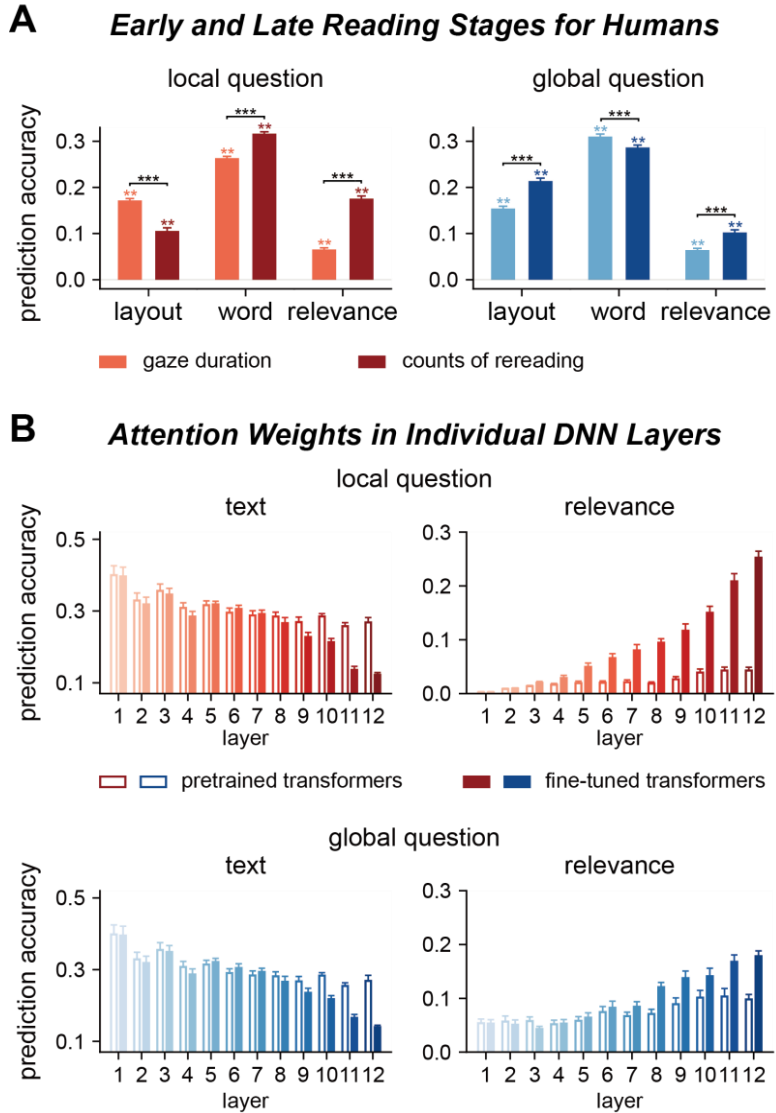

**Fig. 4. Factors influencing attention distribution in different processing stages for humans and DNNs.** (**A**) Human attention in early and late reading stages is differentially modulated by text features and question relevance. The early and late stages are separately characterized by gaze duration, i.e., duration for the first reading of a word, and counts of rereading, respectively. **$P < 0.01$; ***$P < 0.001$. (**B**) DNN attention weights in different layers are also differentially modulated by text features and question relevance. Each attention head is separately modeled and averaged within each layer, and the results are further averaged across the 3 transformer-based models. Shallow layers of both fine-tuned and pre-trained models are more sensitive to text features. Deep layers of fine-tuned models are sensitive to question relevance.

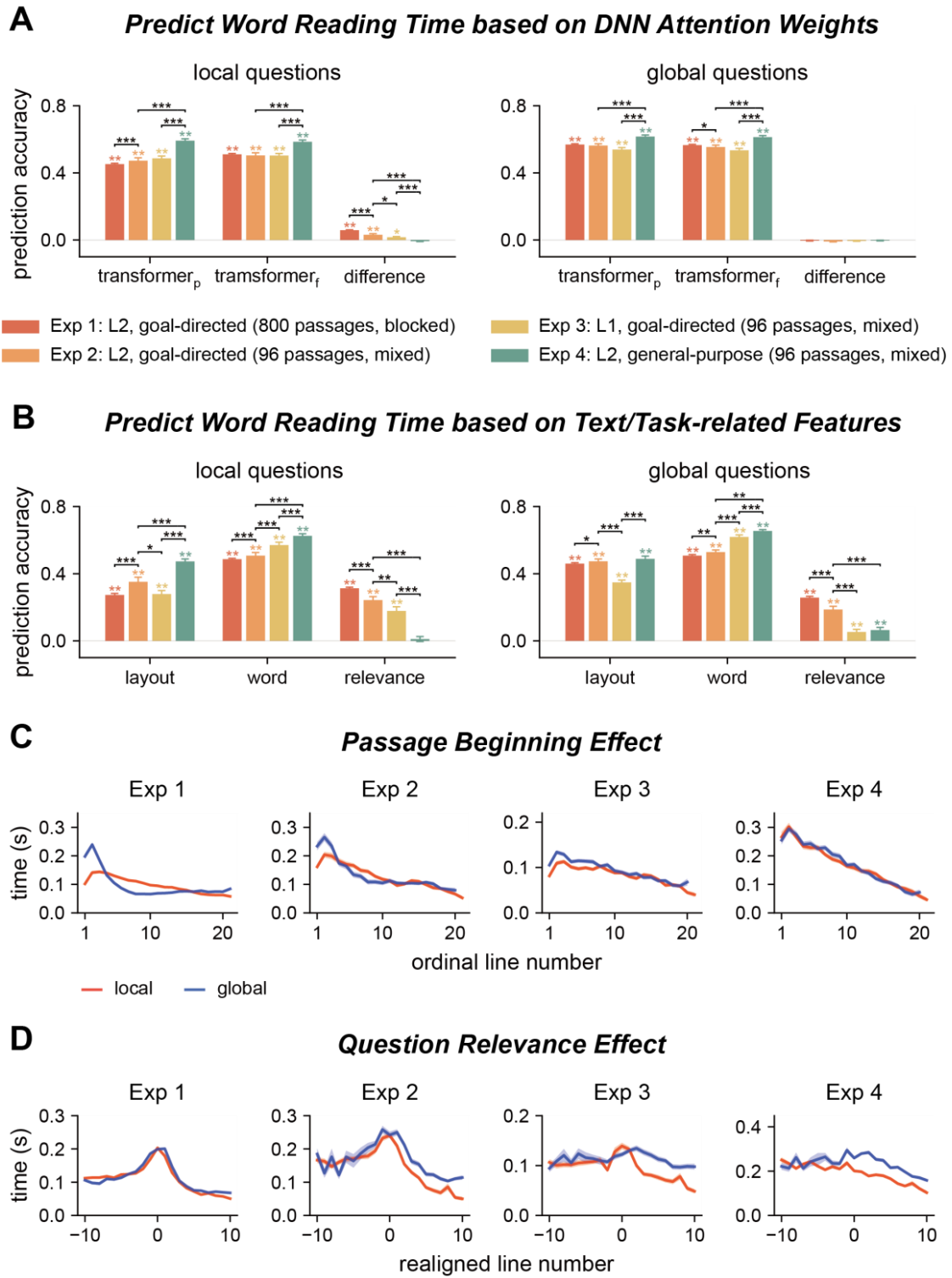

**Fig. 5. Influence of task and language proficiency on word reading time.** (**AB**) Predict the word reading time using attention weights of DNN models, text features, and question relevance for all 4 experiments. Prediction accuracy significantly higher than chance is marked by stars of the same color as the bar. Significant differences

between experiments are denoted by black stars. *P < 0.05; **P < 0.01; ***P < 0.001. (**CD**) Passage beginning and question relevance effects for all 4 experiments. The shade area indicates one SEM across participants.

# Supplementary Material

# Human Attention during Goal-directed Reading Comprehension Relies on Task Optimization

Jiajie Zou, Yuran Zhang, Jialu Li, Xing Tian, Nai Ding[*]

*Corresponding author: Nai Ding

Email: ding_nai@zju.edu.cn

This PDF file includes:

Figures S1 to S8

Tables S1 to S6

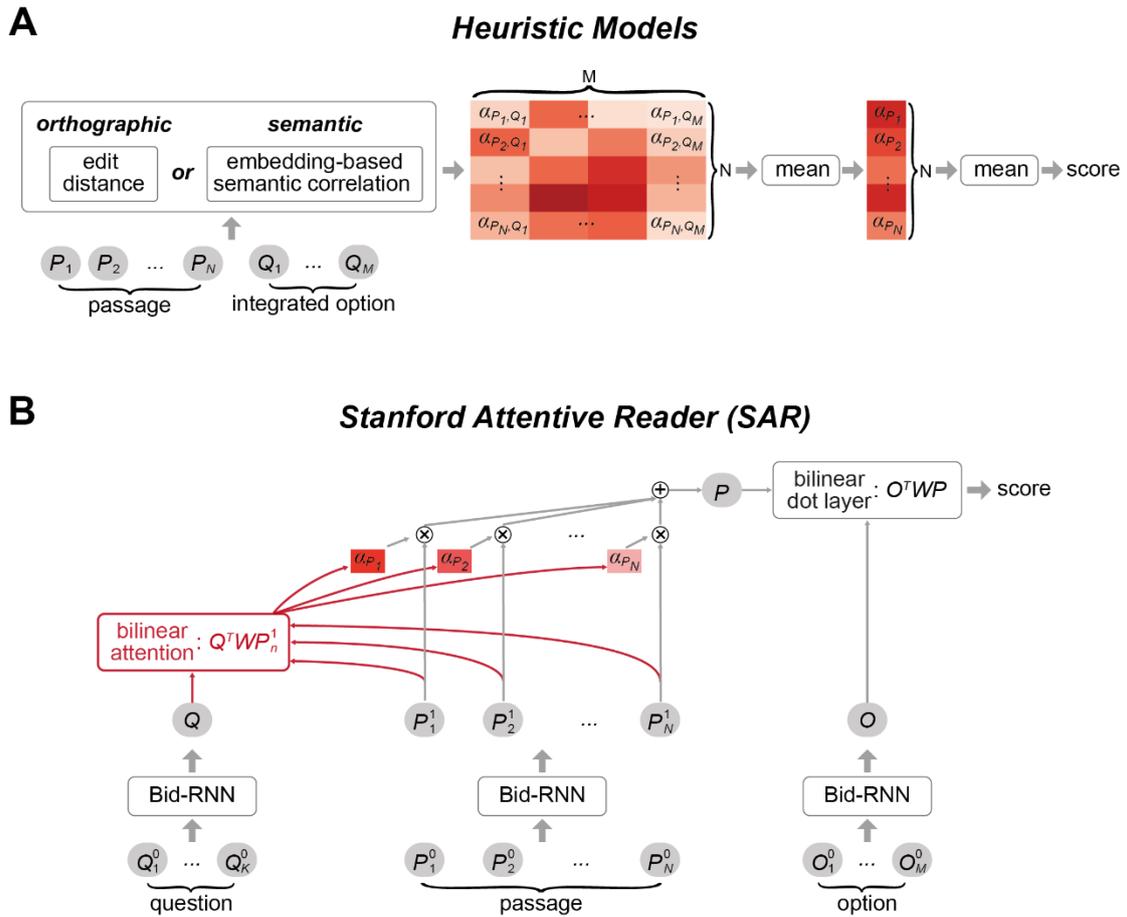

**Fig. S1. Illustration of the word-level heuristic models and the RNN-based SAR model.** (**A**) The orthographic and semantic models calculate the word-wise similarities between all words in the integrated option and all words in the passage, forming a similarity matrix. The similarity measures used in the orthographic and semantic models are the edit distance and correlation between word embeddings, respectively. For each option, the similarity matrix is averaged across all rows and all columns to form a scalar decision score. The option with the largest decision score is chosen as the answer. (**B**) The SAR model uses bi-directional RNNs to encode contextual information. A vectorial representation for the passage is created using the weighted sum of the vectorial representation of each word, and the weight on each word, i.e., the attention weight, is calculated based on its similarity to the vectorial representation of the question. The summarized passage representation and the option representation is used to form the decision score with a bilinear dot layer.

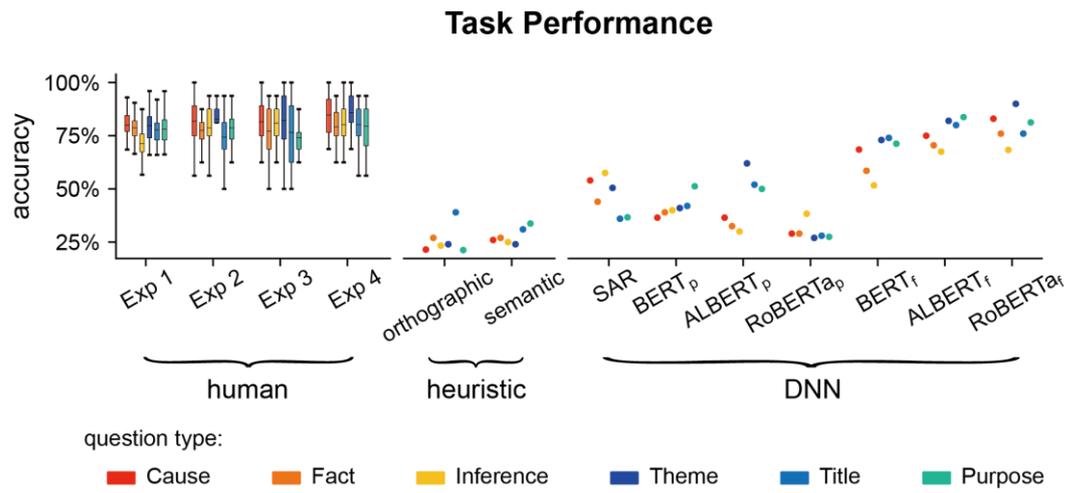

**Fig. S2. Question answering accuracy for individual transformer-based models.**
Human results and other computational models are also plotted for comparison.

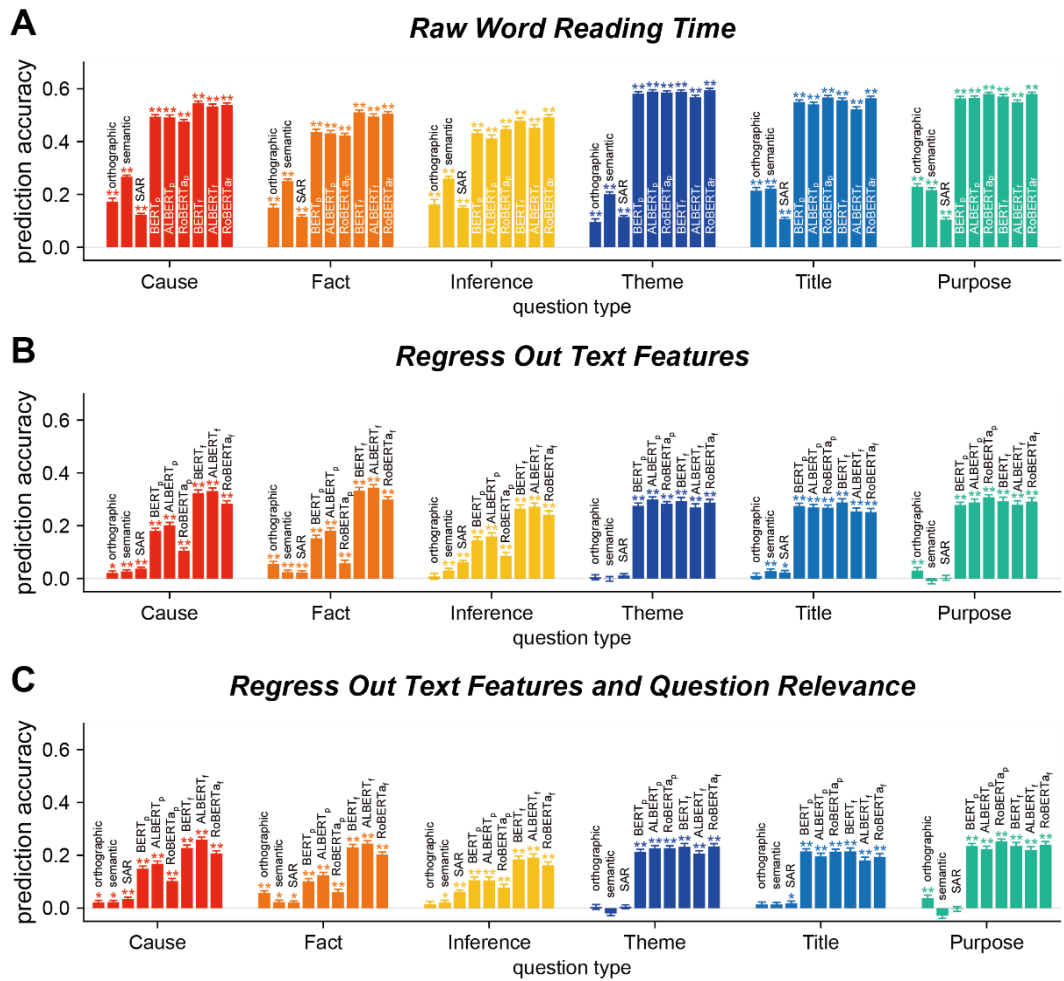

**Fig. S3. Transformer-based models can explain word reading time even when the influences of text features and question relevance are regressed out.** (**A**) Predict the raw word reading time using the attention weights of individual transformer-based models. Results from other computational models are also plotted for comparison. (**B**) Predict the residual word reading time when basic text features, i.e., layout and word features, are regressed out. (C) Predict the residual word reading time when both basic text features and question relevance are regressed out. Prediction accuracy significantly higher than chance is denoted by stars of the same color as the bar. **P < 0.01.

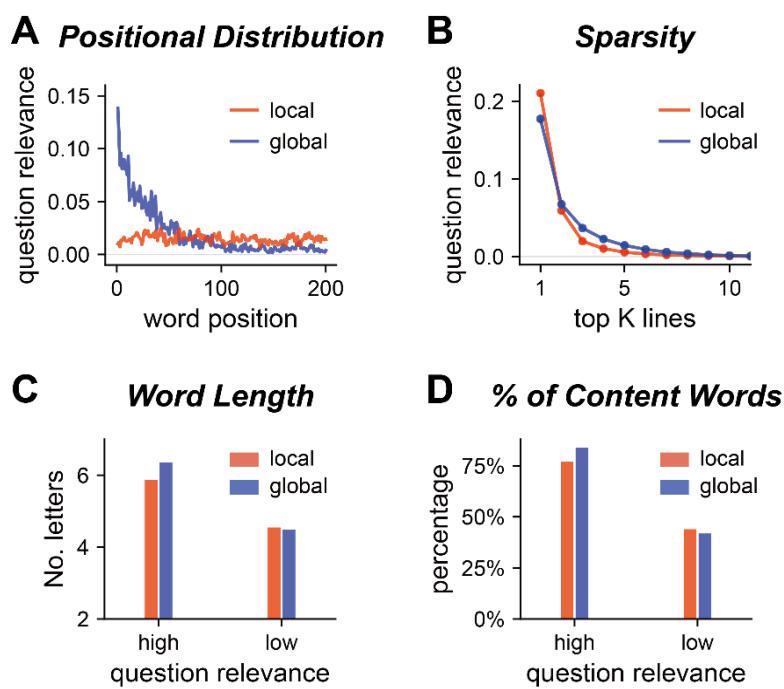

**Fig. S4. Properties of the question relevance of words.** (**A**) Question relevance as a function of the word position within a passage. For global but not local questions, question-relevant words concentrate near the beginning of a passage. (**B**) Decay of mean question relevance across lines. The question relevance is averaged within each line, and all lines in a passage are sorted based on the mean question relevance in descending order. Therefore, line 1 is the line with the highest question relevance, and line 2 is the line with the 2$^{nd}$ highest question relevance. For both global and local questions, the mean question relevance sharply decreases over lines. (**C**) The mean word length, in terms of the number of letters, for words with the question relevance greater or smaller than 0.1. Words of higher relevance are generally longer. (**D**) Percentage of content words for words with higher or lower question relevance. Question-relevant words are more often content words.

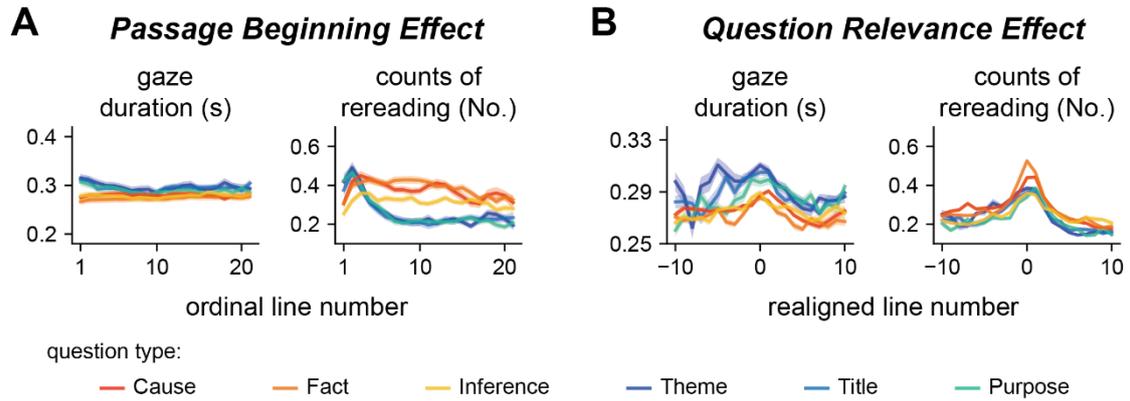

**Fig. S5.** Passage beginning effects (A) and question relevance effects (B) in early and late reading stages. The passage beginning effect differ between global and local questions mainly in the late reading stage reflected by the counts of rereading. The question-relevance effect is also only reliably observed in the late reading stage.

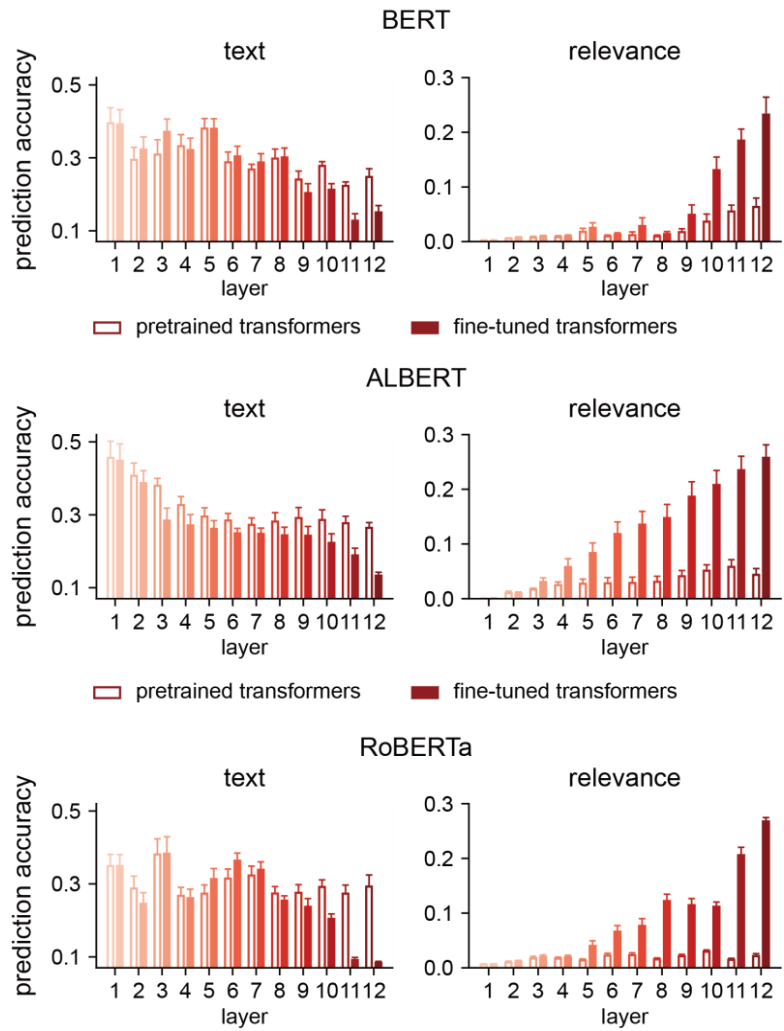

**Fig. S6. Factors influencing attention weights in each layer of DNNs for local questions.** Similar results are observed for all 3 models: The sensitivity to text features decreases from shallow to deep layers, while the sensitivity to question relevance increases across layers.

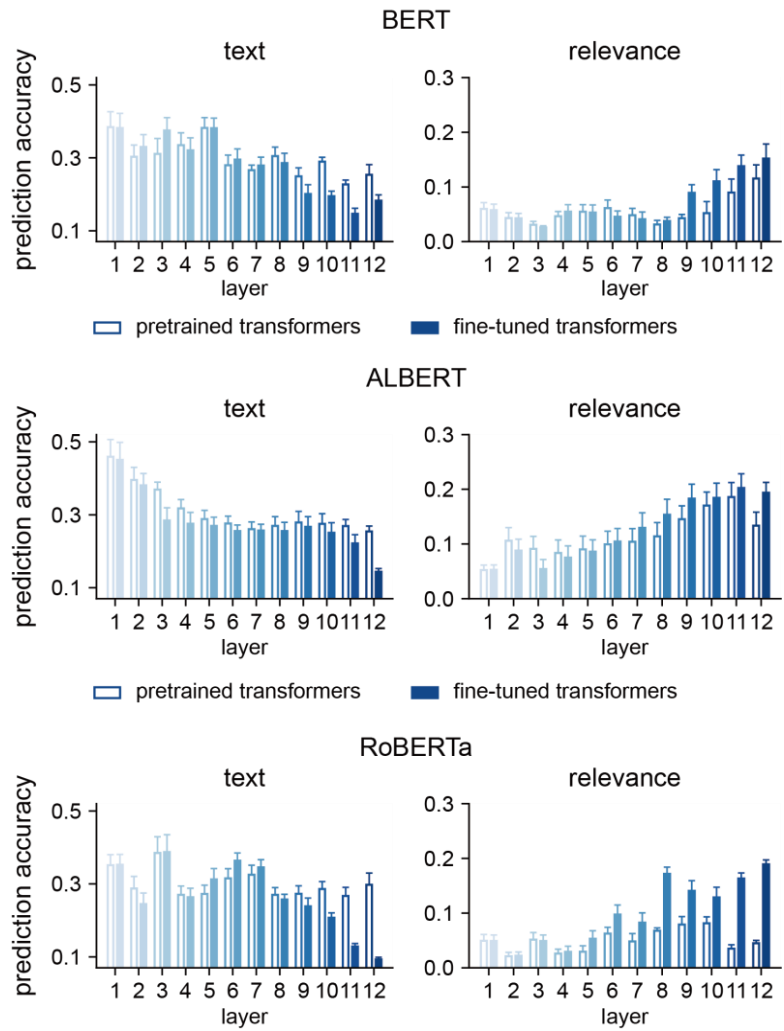

**Fig. S7. Factors influencing attention weights in each layer of DNNs for global questions.** Similar results are observed for all 3 models: The sensitivity to text features decreases from shallow to deep layers, while the sensitivity to question relevance increases across layers.

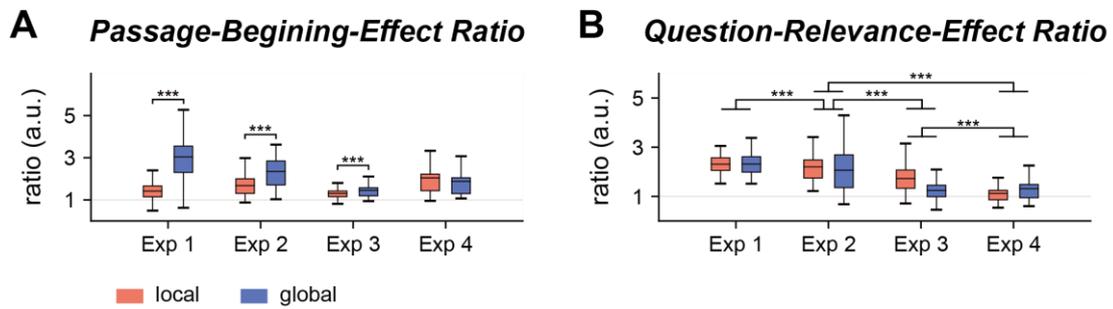

**Fig. S8. Passage-beginning and question-relevance effects in 4 experiments.** The passage beginning effect was quantified by the ratio between the mean word reading time on the first 3 lines of a passage and the mean word reading time on other lines. The question relevance effect was quantified by the ratio between mean word reading time on the line that was most relevant to the question and lines that were more than 5 lines away. See Fig. 1 for the explanation for the box plots. **$P < 0.01$; ***$P < 0.001$.

**Table S1. P-values for the model prediction of word reading time.**

|  | Cause | Fact | Inference | Theme | Title | Purpose |
|---|---|---|---|---|---|---|
| SAR | 0.002 | 0.002 | 0.002 | 0.002 | 0.002 | 0.002 |
| $Trans_p$ | 0.002 | 0.002 | 0.002 | 0.002 | 0.002 | 0.002 |
| $Trans_f$ | 0.002 | 0.002 | 0.002 | 0.002 | 0.002 | 0.002 |
| SAR vs $Trans_p$ | $9\times10^{-5}$ | $9\times10^{-5}$ | $9\times10^{-5}$ | $9\times10^{-5}$ | $9\times10^{-5}$ | $9\times10^{-5}$ |
| SAR vs $Trans_f$ | $9\times10^{-5}$ | $9\times10^{-5}$ | $9\times10^{-5}$ | $9\times10^{-5}$ | $9\times10^{-5}$ | $9\times10^{-5}$ |
| $Trans_p$ vs $Trans_f$ | $9\times10^{-5}$ | $9\times10^{-5}$ | $2\times10^{-4}$ | 0.743 | 0.526 | 0.66 |

*Note*: $Trans_p$: pre-trained transformer-based model; $Trans_f$: fine-tuned transformer-based model.

**Table S2. P-values for the prediction of word reading time using text or task-related features.**

|  | Cause | Fact | Inference | Theme | Title | Purpose |
|---|---|---|---|---|---|---|
| layout | 0.002 | 0.002 | 0.002 | 0.002 | 0.002 | 0.002 |
| word | 0.002 | 0.002 | 0.002 | 0.002 | 0.002 | 0.002 |
| relevance | 0.002 | 0.002 | 0.002 | 0.002 | 0.002 | 0.002 |

**Table S3. P-values for the prediction of early and late eye tracking measures using text or task-related features.**

|  | (i) layout | | | (ii) word | | (iii) relevance | |
|---|---|---|---|---|---|---|---|
|  | local | global | local vs global | local | global | local | global |
| GD | 0.003 | 0.003 | 1 | 0.002 | 0.002 | 0.002 | 0.002 |
| CR | 0.003 | 0.003 | $1\times10^{-4}$ | 0.002 | 0.002 | 0.002 | 0.002 |

*Note*: GD: gaze duration; CR: counts of rereading.

**Table S4. P-values for the prediction of word reading time for all 4 experiments.**

|  | local questions | | | | global questions | | | |
|---|---|---|---|---|---|---|---|---|
|  | Ex 1 | Ex 2 | Ex 3 | Ex 4 | Ex 1 | Ex 2 | Ex 3 | Ex 4 |
| $Trans_p$ | 0.003 | 0.003 | 0.003 | 0.003 | 0.004 | 0.004 | 0.004 | 0.004 |
| $Trans_f$ | 0.003 | 0.003 | 0.003 | 0.003 | 0.004 | 0.004 | 0.004 | 0.004 |
| $Trans_f$ - $Trans_p$ | 0.003 | 0.003 | 0.041 | 0.814 | 0.9 | 0.9 | 0.829 | 0.714 |
| layout | 0.002 | 0.002 | 0.002 | 0.002 | 0.003 | 0.003 | 0.003 | 0.003 |
| word | 0.002 | 0.002 | 0.002 | 0.002 | 0.003 | 0.003 | 0.003 | 0.003 |
| relevance | 0.002 | 0.002 | 0.002 | 0.228 | 0.003 | 0.003 | 0.003 | 0.003 |

**Table S5. P-values for the comparisons between experiments.**

|  | local questions | | | | global questions | | | |
|---|---|---|---|---|---|---|---|---|
|  | Ex 1 vs Ex 2 | Ex 2 vs Ex 3 | Ex 2 vs Ex 4 | Ex 3 vs Ex 4 | Ex 1 vs Ex 2 | Ex 2 vs Ex 3 | Ex 2 vs Ex 4 | Ex 3 vs Ex 4 |
| $Trans_p$ | 0.001 | 0.460 | $1\times10^{-4}$ | $1\times10^{-4}$ | 0.155 | 0.055 | $1\times10^{-4}$ | $1\times10^{-4}$ |
| $Trans_f$ | 0.2 | 0.953 | $1\times10^{-4}$ | $1\times10^{-4}$ | 0.02 | 0.114 | $1\times10^{-4}$ | $1\times10^{-4}$ |
| $Trans_f$ - $Trans_p$ | $1\times10^{-4}$ | 0.032 | $1\times10^{-4}$ | $1\times10^{-4}$ | 0.04 | 0.305 | 0.12 | 0.622 |
| layout | $1\times10^{-4}$ | 0.014 | $1\times10^{-4}$ | $1\times10^{-4}$ | 0.021 | $1\times10^{-4}$ | 0.307 | $1\times10^{-4}$ |
| word | 0.001 | 0.001 | $1\times10^{-4}$ | 0.001 | 0.003 | $1\times10^{-4}$ | $1\times10^{-4}$ | 0.003 |
| relevance | $1\times10^{-4}$ | 0.002 | $1\times10^{-4}$ | $1\times10^{-4}$ | $1\times10^{-4}$ | $1\times10^{-4}$ | $1\times10^{-4}$ | 0.496 |

**Table S6. Hyperparameters for DNN fine-tuning.** We adapted these hyperparamemers from references [1-4].

| models | learning rate | fine-tuning steps | fine-tuning batch size | warmup steps | weight decay |
|---|---|---|---|---|---|
| BERT | $1\times10^{-5}$ | 27455 | 16 | 0 | 0 |
| ALBERT | $2\times10^{-5}$ | 12000 | 32 | 1000 | 0 |
| RoBERTa | $1\times10^{-5}$ | 21964 | 16 | 1200 | 0.1 |

**References**

1. Lan Z, Chen M, Goodman S, Gimpel K, Sharma P, Soricut R. Albert: A lite bert for self-supervised learning of language representations. International Conference on Learning Representations; 2019: ICLR; 2020.
2. Liu Y, Ott M, Goyal N, Du J, Joshi M, Chen D, et al. Roberta: A robustly optimized bert pretraining approach. arXiv. 2019. doi: arXiv:1907.11692.
3. Zhang S, Zhao H, Wu Y, Zhang Z, Zhou X, Zhou X. DCMN+: Dual co-matching network for multi-choice reading comprehension. AAAI conference on artificial intelligence; 2019: AAAI 2020.
4. Ran Q, Li P, Hu W, Zhou J. Option comparison network for multiple-choice reading comprehension. arXiv. 2019. doi: arXiv:1903.03033.